\pdfoutput=1

\documentclass[11pt]{article}

\usepackage[final]{acl}

\usepackage{amsmath,amsfonts,bm}

\newcommand{\policy}{\pi}
\newcommand{\reference}{\pi_\text{ref}}

\newcommand{\newterm}[1]{{\bf #1}}

\def\eqref#1{equation~\ref{#1}}

\def\1{\bm{1}}
\newcommand{\train}{\mathcal{D}}

\DeclareMathAlphabet{\mathsfit}{\encodingdefault}{\sfdefault}{m}{sl}
\SetMathAlphabet{\mathsfit}{bold}{\encodingdefault}{\sfdefault}{bx}{n}

\newcommand{\E}{\mathbb{E}}
\newcommand{\Ls}{\mathcal{L}}

\newcommand{\KL}{D_{\mathrm{KL}}}

\usepackage{times}
\usepackage{latexsym}

\usepackage[T1]{fontenc}

\usepackage[utf8]{inputenc}

\usepackage{microtype}

\usepackage{inconsolata}

\usepackage{graphicx}
\usepackage{amsmath}
\usepackage{xspace}
\usepackage{subcaption}
\usepackage{array}
\usepackage{wrapfig}
\usepackage{url}
\usepackage{tcolorbox}
\usepackage{booktabs}
\usepackage{amssymb}
\usepackage{bbding}
\usepackage{xcolor}
\usepackage{makecell}
\usepackage{tikz}
\usepackage{pifont}
\usetikzlibrary{shapes}
\usepackage{multirow}
\usepackage{minitoc}
\usepackage{listings}
\usepackage{graphicx}
\usepackage{pifont}
\usepackage{algpseudocode}
\usepackage[linesnumbered,ruled,vlined]{algorithm2e}
\SetKwInOut{KwData}{Input}
\SetKwInOut{KwResult}{Output}
\usepackage{enumitem}

\newcommand{\OURS}{\text{Delta-KD}\xspace}

\title{Delta Knowledge Distillation for Large Language Models}

\author{%
  Yihan Cao \quad Yanbin Kang \quad Zhengming Xing \quad Ruijie Jiang\thanks{Corresponding author.} \\  
  LinkedIn Corporation \\[0.5ex]
  \texttt{\{yihacao,ybkang,zhxing,rjiang\}@linkedin.com} \\
}

\begin{document}
\maketitle
\begin{abstract}
Knowledge distillation (KD) is a widely adopted approach for compressing large neural networks by transferring knowledge from a large teacher model to a smaller student model. In the context of large language models, token-level KD, typically minimizing the KL divergence between student output distribution $\pi_s$ and teacher output distribution $\pi_t$, has shown strong empirical performance. 
However, prior work typically assumes $\pi_t$ and $\pi_s$ share the same optimal representation space — a premise that may not hold in many cases. To solve this problem, 
we propose Delta Knowledge Distillation (\OURS), a novel extension of token-level KD that encourages the student to approximate an optimal representation space by explicitly preserving the distributional shift $\Delta$ introduced during the teacher’s supervised finetuning (SFT). 
Empirical results on ROUGE metrics demonstrate that \OURS~substantially improves student performance while preserving more of the teacher’s knowledge.\footnote{Work in progress.}

\end{abstract}


\section{Introduction}
Large language models (LLMs) have demonstrated remarkable performance across a wide range of natural language processing (NLP) tasks~\cite{tie2025large, minaee2024large}. Their success can be attributed to their massive scale, which enables them to learn world knowledge and following human instructions. However, the computational cost of deploying these models is substantial, making them impractical for many resource-constrained applications. In such scenarios, smaller models are preferred due to their lower latency and memory footprint. This motivates the need for model compression techniques that can retain the capabilities of LLMs in smaller architectures.

\begin{figure}[ht]
\centering
\includegraphics[width=1\linewidth]{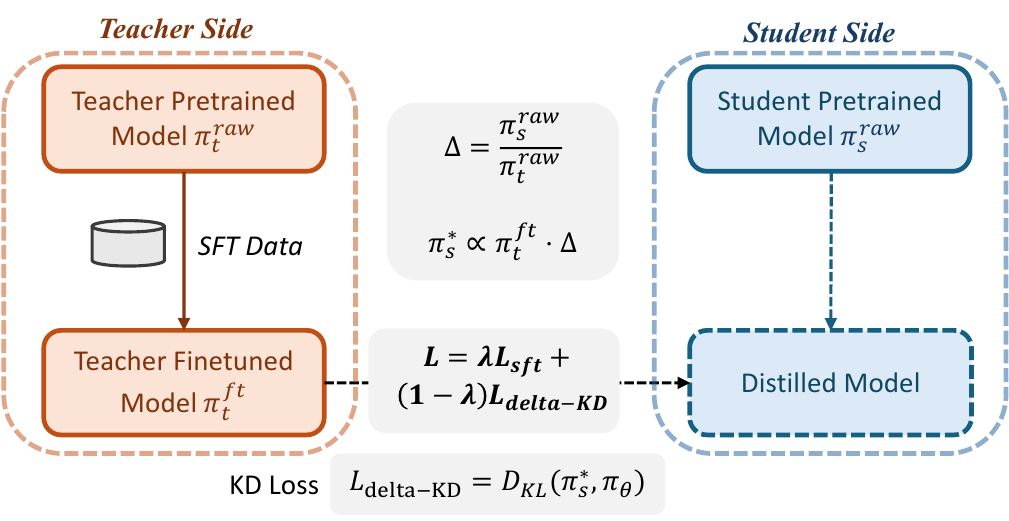}
\caption{
\OURS overview.
Orange boxes refer to the teacher-side components, where the pretrained teacher model $\pi_t^{\text{raw}}$ is finetuned on SFT data to obtain $\pi_t^{\text{ft}}$. Blue boxes refer to the student-side components, where the pretrained student model $\pi_s^{\text{raw}}$ is distilled into the final model. Grey boxes denote the core computations in \OURS: the distributional shift $\Delta$, the target distribution $\pi_s^*$, and the final loss $\mathcal{L}$, which combines supervised finetuning and the proposed \OURS loss.
}
\label{example_abs}
\vspace{-8pt}
\end{figure} 

One widely adopted solution is knowledge distillation (KD), a technique originally developed in the computer vision domain and now extensively used for compressing large language models~\cite{xu2024survey}. In KD, a smaller student model is trained to replicate the behavior of a larger teacher model, enabling significant reductions in serving cost without substantial loss in performance. Specifically, KD in large language models can be applied in either \textit{black-box} or \textit{white-box} settings~\cite{xu2024survey}. In the \textit{black-box} scenario, the teacher model is treated as an API: it is prompted to generate output sequences, which are then used to finetune the student via supervised learning. In contrast, \textit{white-box} approaches leverage full access to the teacher’s internal outputs. Token-level KD minimizes the KL divergence between the teacher and student logits, while sequence-level KD aligns the distributions over sampled sequences from each model~\cite{cao2025llm}. These methods allow for finer-grained supervision and more effective knowledge transfer.

However, despite their effectiveness, existing knowledge distillation methods may suffer from two key limitations. 
First, prior approaches assume that student and teacher models operate in the same representation space, directly minimizing the distance between their output distributions. This assumption may not hold, especially when compressing to much smaller models, leading to suboptimal alignment and degraded performance.
Second, prior methods typically overlook the rich knowledge encoded in the pretrained model and fail to capture the distribution shift that occurs during supervised finetuning. By relying solely on the final finetuned teacher, these methods discard potentially valuable intermediate signals. 

To tackle with these problems, we propose \OURS, a novel approach for token-level KD that encourages approximation of the optimal student representation space via maintaining the distribution shift, instead of directly aligns the student and teacher output distributions. 
Specifically, we start with defining the $\Delta$ term. 
Given a small student language model $\pi_s$ and a large teacher language model $\pi_t$, for a given instruction $x$, the output distribution over response $y$ is represented by $\pi_s(y|x)$ for the student LM and $\pi_t(y|x)$ for the teacher LM. We define the delta term as $\Delta_{s\rightarrow t}=\frac{\pi_s}{\pi_t}$ which serves as an implicit reward signal~\cite{mitchell2023emulator}. 
This term captures the relative preference of the teacher model's preference over the student model~\cite{mitchell2023emulator}, effectively quantifying how much more likely the large model favors a response compared to the student model output. 

In KD, let $\pi_s$ denote a student model distribution, and let $\pi_t^{\text{raw}}$ and $\pi_t^{\text{ft}}$ represent the teacher model before and after finetuning, respectively. We define the synthetic output distribution as $\pi_s^* = \Delta_t^{\text{raw} \rightarrow \text{ft}} \cdot \pi$, where $\Delta_t^{\text{raw} \rightarrow \text{ft}}$ captures the knowledge shift from the teacher's pretraining to finetuning. Rather than directly aligning $\pi_s$ with $\pi_t^{\text{ft}}$, we align $\pi_s$ with $\pi_s^*$. This approach enriches the distillation process by incorporating both pretraining and finetuning signals from the teacher, while maintaining alignment within the student’s own representation space.

We conduct both token-level and sequence-level knowledge distillation experiments using the widely adopted \texttt{Qwen2.5} model series on two datasets: an instruction tuning dataset (\texttt{ultrachat-200k}~\cite{ding2023enhancing}) and a reasoning dataset (\texttt{OpenMathReasoning}~\cite{moshkov2025aimo2}).
Model performance is evaluated using ROUGE scores mainly and will be evaluated using feedback from strong LLMs in the future.
Extensive experiments demonstrate that \OURS consistently outperforms state-of-the-art KD baselines. Our contributions are summarized as follows:

\begin{itemize}[leftmargin=*]
    \item We propose a novel language model knowledge distillation method, \OURS, which emulates the behavior of a finetuned teacher by incorporating the distribution shift learned during teacher's SFT stage into the distillation process.
    \item We decouple teacher inference from student training via ZeroMQ communication and Python shared memory, enabling delta-distillation with long sequences and large models under GPU memory constraints.
    \item We conduct extensive token-level and sequence-level distillation experiments across datasets in different domains. Empirical results on related evaluation  metrics demonstrate that \OURS outperforms strong baselines.
    \item We further introduce a \textit{Parallelogram Loss} to extend our framework, enabling effective combinations of distillation signals from multiple teacher variants along different paths in the teacher model space.
\end{itemize}

\section{Knowledge Distillation Setup}
\label{sec:distill-setup}

We consider the standard setting of large language model (LLM) distillation, 
in which the goal is to transfer task-specific behavior from a large, 
computationally expensive \newterm{teacher model} to a smaller, more 
efficient \newterm{student model}. This setup is motivated by the 
observation that while large models (e.g., \texttt{LLaMA-65B}) typically 
achieve superior performance, they are often impractical to deploy due to 
resource constraints. Distillation offers a way to preserve much of the 
teacher's output quality in a lightweight model that is cheaper to run 
at inference time.

Let \( x \in \mathcal{X} \) denote an input prompt, such as an instruction 
or natural language query, and \( y \in \mathcal{Y} \) denote a textual 
response. Both teacher and student models define conditional distributions 
over \( y \) given \( x \), parameterized by policies 
\( \reference(y \mid x) \) and \( \policy_{\theta}(y \mid x) \), respectively. 
We assume access to a dataset \( \train = \{x_i\}_{i=1}^N \) of input prompts, 
and optionally paired output responses \( y_i \), if human-labeled 
reference outputs are available.

Our objective is to learn a student policy \(\policy_{\theta}(y \mid x)\) that closely approximates the output behavior of the teacher policy \(\reference(y \mid x)\), while maintaining lower inference cost and parameter count.

To facilitate stable and informative supervision, we apply \newterm{temperature-scaled softmax} to both teacher and student output logits. Let \( z_{\text{ref}}(y \mid x) \) and \( z_{\theta}(y \mid x) \) denote the unnormalized logits output by the teacher and student models, respectively. The corresponding softened distributions are:
\begin{align}
    \pi_{\text{ref}}(y \mid x) 
    &= \frac{e^{ z_{\text{ref}}(y \mid x) / \tau}}{
        \sum_{y' \in \mathcal{Y}} e^{ z_{\text{ref}}(y' \mid x) / \tau }}, \\
    \policy_{\theta}(y \mid x) 
    &= \frac{e^{ z_{\theta}(y \mid x) / \tau }}{
        \sum_{y' \in \mathcal{Y}} e^ {z _{\theta}(y' \mid x) / \tau}},
\end{align}
where \( \tau > 0 \) is a temperature parameter controlling the sharpness 
of the output distribution. Higher \( \tau \) values result in softer 
(probabilistically flatter) distributions, allowing the student to learn 
from the full probability mass rather than only the top-1 token.

Our primary distillation loss is the token-level Kullback–Leibler divergence 
between the softened teacher and student output distributions:
\begin{equation}
\Ls_{\text{KD}} = \E_{x \sim \train} 
\left[ \KL\left( {\reference}(\cdot \mid x) 
\| {\policy}_{\theta}(\cdot \mid x) \right) \right],
\end{equation}
which encourages the student to match the entire output distribution of 
the teacher, not just the most likely token.

When ground-truth targets \( y \sim \mathcal{Y} \) are available, we 
also include a supervised learning term that trains the student to 
maximize the likelihood of correct outputs:
\begin{equation}
\Ls_{\text{SFT}} = - \E_{(x, y) \sim \train} 
\left[ \log \policy_{\theta}(y \mid x) \right].
\end{equation}

The total training objective for the student model combines these two 
objectives:
\begin{equation}
\Ls_{\text{total}} = \lambda \Ls_{\text{SFT}} 
+ (1 - \lambda) \Ls_{\text{KD}},
\end{equation}
where \( \lambda \in [0, 1] \) is a tunable hyperparameter that balances 
the importance of exact supervision versus imitation of the teacher.

This distillation framework supports a wide range of LLM training regimes, 
including instruction tuning, chat modeling, summarization, and 
preference alignment. The KL term transfers knowledge from the 
teacher’s output distribution, encoding both semantic fluency and 
task-specific reasoning. The optional supervised loss ensures that the 
student retains grounding in human-labeled correctness when available. 
In practice, this approach enables highly efficient deployment of 
LLMs without the need to finetune large models directly. In this paper we ask: is there a better way to choose a teacher model?

\section{Delta Knowledge Distillation}
In this section we describe our approach for teacher model design. We begin by asking what makes a good teacher model? To answer this question we adopt the following two guiding principles:

\textbf{Principle 1.} A student model should not be forced to exactly match the teacher’s output distribution, as the representational capacities of large and small models are fundamentally different.

\textbf{Principle 2.} Instead of mimicking absolute outputs, the student should aim to replicate the teacher's \textit{behavioral shift} from pretraining to finetuning—that is, the transformation induced 
by alignment should be consistent across model scales.

In short, we argue that the most effective distillation signal is not the absolute output of the teacher, but rather the direction in which the teacher deviates from the base model during finetuning. By focusing on this behavioral delta, we allow the student to learn scale-appropriate task-specific knowledge that generalizes more robustly. This setup naturally satisfies both of our principles: it avoids forcing the student to match unreachable outputs, and it preserves the finetuning trajectory that aligns with the task objective.

In practice, however, Principle 2 cannot be directly implemented, as we do not have access to a finetuned model of the same capacity as the student. To address this challenge, we propose a method that 
does not require learning directly from the large teacher model (consistent with Principle 1), but instead approximates the behavioral shift of a hypothetical same-scale finetuned student.

Specifically, we introduce a parameterized teacher distribution that interpolates between the student’s raw output and the large model’s finetuned output. This interpolation is controlled by a scalar 
parameter \( \alpha \in [0, 1] \), which smoothly adjusts the degree of behavioral shift supervision. When \( \alpha = 0 \), the teacher reduces to the raw student model, and when \( \alpha = 1 \), it fully 
reflects the behavior delta derived from the large model. 

This flexibility is critical: allowing either component to dominate can harm performance—naively copying the large model overwhelms the student's capacity, while relying solely on the student’s own prior yields under-adaptation. Our method finds a balance that generalizes more effectively across model scales.

\subsection{Proposed LLM Distillation}
\label{sec:proposed-distill}

Our goal is to construct an effective teacher distribution $\reference(y \mid x)$ for training a small student model $\policy_{\theta}(y \mid x)$. This teacher distribution should incorporate information from both the pretrained and finetuned behaviors of large and small models. Specifically, we have access to the following fixed, non-learnable distributions:

{\small
\begin{itemize}
\item $\pi_{s}^{\text{raw}}(y \mid x)$: the small model before finetuning;
\item $\pi_{t}^{\text{raw}}(y \mid x)$: the large model before finetuning;
\item $\pi_{t}^{\text{ft}}(y \mid x)$: the large model after finetuning.
\end{itemize}
}

Our objective is to use these distributions to approximate the following latent optimal target:

{\small
\begin{itemize}
\item $\pi_{s}^{*}(y \mid x)$: the (unobserved) optimal distribution that the small model should represent after finetuning.
\end{itemize}
}

Since we do not have access to $\pi_{s}^{*}(y \mid x)$, we propose a method to approximate it using the available teacher distributions. This approximation directly defines our teacher distribution:

$$
\reference(y \mid x) = \pi_{s}^{*}(y \mid x).
$$

Note that at the start of training, the student model  $\pi_{\theta}(y \mid x)$ is initialized with the parameters of $\pi_{s}^{\text{raw}}(y \mid x)$. However, the distributions $\pi_{s}^{\text{raw}}(y \mid x)$, $\pi_{t}^{\text{raw}}(y \mid x)$, and $\pi_{t}^{\text{ft}}(y \mid x)$ remain fixed and do not change during training.

According to our design principles, we aim to align the 
finetuning trajectory of the small model with that of the 
large model by matching their respective behavior shifts. 
To formalize this, we define a generic shift operator 
\(\Delta(p_1, p_2)(y \mid x) := \frac{p_1(y \mid x)}{p_2(y \mid x)}\), 
which captures the relative change from a base distribution \(p_2\) 
to a target distribution \(p_1\).

We posit that the shift experienced by the small model should align 
with the shift observed in the large model, up to a scale factor. 
Specifically, we require:
\begin{equation}
    \Delta(\pi_{s}^{*}, \pi_{t}^{\text{ft}})(y \mid x) 
    \propto 
    \Delta(\pi_{s}^{\text{raw}}, \pi_{t}^{\text{raw}})(y \mid x)^{\alpha},
    \label{eq:delta-alignment}
\end{equation}


where \( \pi_{s}^{*}\) is the target student 
distribution we aim to construct, and \(\alpha \) controls the 
intensity of alignment. This formulation encourages the student model 
to follow the same transformation pattern that the large model 
undergoes during finetuning, while respecting the student’s 
limited capacity.

To incorporate this alignment principle into the student training objective, 
we construct a synthetic teacher distribution \(\pi_{s}^{*}(y \mid x)\) 
that satisfies the proportional relationship in Eq.~\ref{eq:delta-alignment}. 
Concretely, we insert this target distribution into a KL-based distillation loss:
\begin{equation}
    \Ls_{\text{delta-KD}} = \E_{x \sim \train}
    \left[ \KL\left(\pi_{s}^{*}(\cdot \mid x) 
    \;\|\; \pi_{\theta}(\cdot \mid x) \right) \right].
\end{equation}

The only requirement to make Eq.~\ref{eq:delta-alignment} a valid probability distribution is to normalize its right-hand side. Specifically, we define \(\pi_{s}^{*}(y_t \mid x)\) as:
{\small
\begin{equation}
     \pi_{s}^{*}(y_t \mid x) = 
    \frac{1}{Z(x, y)} 
    \cdot  \pi_{t}^{\text{ft}}(y_t \mid x)
    \cdot (\frac{\pi_{s}^{\text{raw}}(y_t \mid x)  }{ \pi_{t}^{\text{raw}}(y_t \mid x) })^{\alpha},
\end{equation}
\label{eq:delta-eq-8}
}
where \( Z(x, y) \) is a token-level partition function ensuring normalization:

\begin{equation}
    Z(x, y) = \sum_{y_t \in \mathcal{V}} 
      \pi_{t}^{\text{ft}}(y_t \mid x)
    \cdot (\frac{\pi_{s}^{\text{raw}}(y_t \mid x) }{ \pi_{t}^{\text{raw}}(y_t \mid x) })^{\alpha}.
\end{equation}

This formulation is analogous to prior work in preference modeling 
\citep{mitchell2023emulator}, where the reward function is reinterpreted 
as an \textit{advantage function}. In our case, the delta-guided 
teacher distribution reflects a scaled advantage over the base 
(student raw) distribution, encoding how the teacher's behavior 
changes due to finetuning.

By distilling from this normalized form, we ensure that the student 
model learns not to reproduce the teacher’s absolute behavior, but 
rather to replicate the \textit{change in behavior}—modulated 
appropriately to respect the capacity and inductive bias of the student.

While delta distillation provides valuable behavioral alignment, it 
may not fully capture the task-specific signal encoded in supervised 
finetuning (SFT) data. To address this, we integrate both sources 
of supervision by combining the delta-based distillation loss with 
a conventional SFT loss (e.g., cross-entropy on labeled responses). 
The final training objective is a weighted combination:

\begin{equation}
\Ls_{\text{total}} = \lambda \Ls_{\text{SFT}} 
+ (1 - \lambda) \Ls_{\text{delta-KD}},
\end{equation}

where \(\lambda \in [0, 1]\) controls the trade-off between direct 
supervised supervision and delta-aligned imitation learning.

\newpage

\section{Experiment}

\begin{figure}[t]
\centering
\includegraphics[width=1\linewidth]{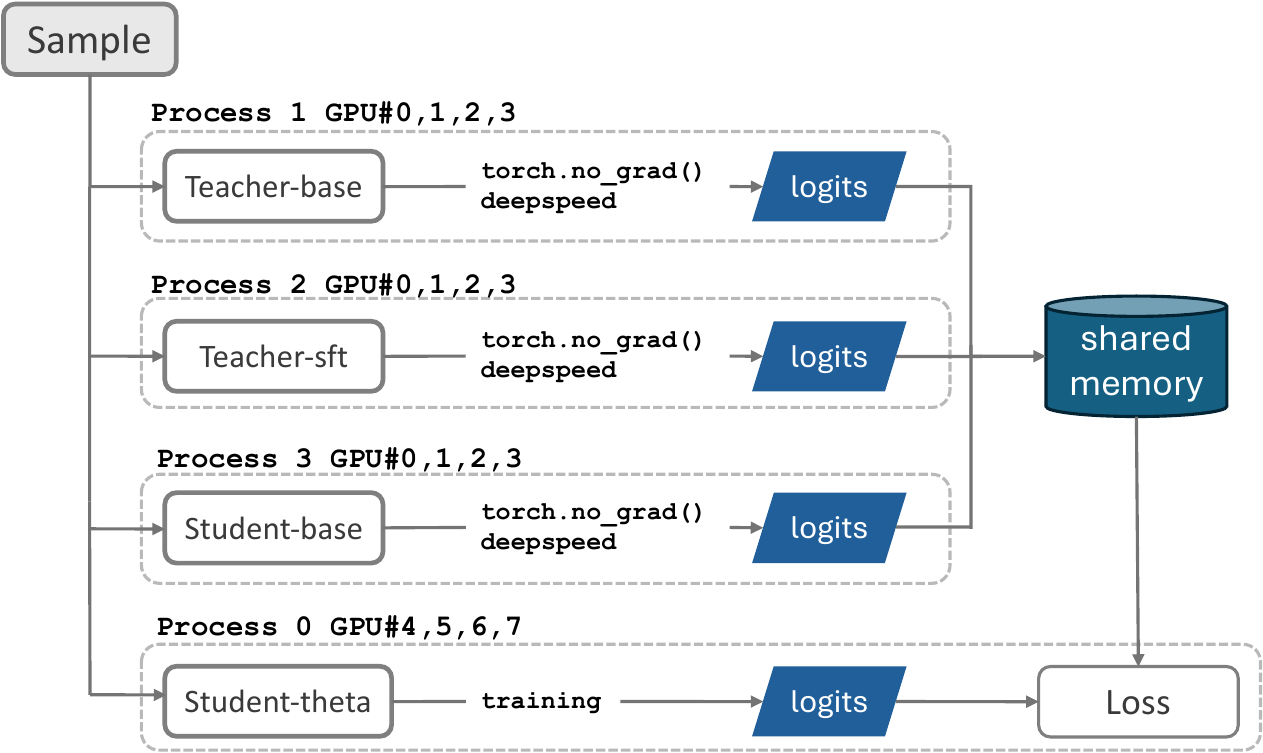}
\caption{Illustration of using ZeroMQ during training.
Grey boxes refer to different model variants used during distillation, including pretrained teacher model (teacher-base), finetuned teacher model (teacher-sft), student model (student-base) and distilled student model (student-theta). Blue boxes denote the generated logits from each model, which are stored in shared memory for efficient communication. The upper three processes run in inference mode using DeepSpeed across GPUs \# 0–3, while the bottom process trains the student-theta model on GPUs \# 4–7 by consuming logits from shared memory and computing the final loss. 
}
\label{fig:zeromq}
\vspace{-8pt}
\end{figure} 

\subsection{Implementation Details}
In this section, we present the implementation details, including the training and evaluation settings and distillation training architecture specifically designed for \OURS.
\subsubsection{Training settings.} 
\paragraph{Datasets.}We run our experiments on widely used instruction tuning and reasoning datasets across different base models. Given that LLaMA-3 and Qwen models already achieve strong performance on instruction-following tasks, we use the advanced \texttt{ultrachat-200k} \footnote{https://huggingface.co/datasets/HuggingFaceH4/\\ultrachat\_200k} dataset to assess instruction-following capabilities, and \texttt{OpenMathReasoning} \footnote{https://huggingface.co/datasets/nvidia/OpenMathReasoning} to evaluate reasoning and question answering abilities. To reduce training time, we sampled the original datasets and only do finetuning and distillation on a small subset.

\paragraph{Base models.}For base models, we focus on two widely used model families: \texttt{LLaMA-3} and \texttt{Qwen2.5}. Currently we only present results with \texttt{Qwen2.5} and in the future, we'll provide more results on different base models. In all experiments, 7B-scale models serve as teachers and 1.5B-scale models as students. Specifically, we use \texttt{Qwen-2.5-1.5B} as student models. For teacher models, we use \texttt{Qwen-2.5-7B} as reference models, along with their corresponding finetuned versions for the teacher finetuned models.

\paragraph{Other settings.}Both training and evaluation are conducted on an instance with 8 NVIDIA H100 80GB GPUs. The maximum sequence length is set to 4096 for \texttt{ultrachat-200k} experiments and 8192 for \texttt{OpenMathReasoning}. Batch size is set to 2 and gradient accumulation steps is set to 16.

\subsubsection{Training Architecture}
Training with Eq.~\ref{eq:delta-alignment} requires retrieving the logits output from multiple LLMs during distillation. These models are typically memory intensive, making it infeasible to colocate them with the student model on the same GPU without hitting memory constraints. This significantly limits the size of models and the maximum sequence length that can be used in training.  

Two conventional strategies are commonly adopted to acquire teacher logits during distillation:

\begin{itemize}

 \item \textbf{Offline storage:} Precompute and store teacher logits on disk before training, then load them on the fly. While straightforward, this method becomes impractical when dealing with large models and long sequences, as the disk I/O bottleneck and storage requirements can be prohibitive.

 \item \textbf{On-device inference:} Run both the teacher and student models on the same GPU during training. This approach restricts the scale of training due to the substantial memory footprint of teacher models.

\end{itemize}

These challenges are exacerbated in the \OURS setting, where training requires logits from \textit{two} teacher models—one raw and one supervised finetuned (SFT)—in addition to the student model. Placing all these models on the same GPU is infeasible due to memory limitations.

To address this issue, we decouple inference and training: the teacher models are deployed as standalone services on dedicated GPUs. Figure~\ref{fig:zeromq} shows the training architecture. During training, the student process communicates with the teacher inference processes using \textbf{ZeroMQ} sockets. The input queries are sent via ZeroMQ, and the resulting teacher logits are transferred using \textbf{Python shared memory}, which avoids unnecessary serialization overhead. Furthermore, to reduce communication bandwidth while maintaining acceptable numerical fidelity, all logits are transmitted in \textbf{FP16} format.

This design enables scalable, memory-efficient training and supports long input sequences without compromising on model size.

\subsubsection{Evaluation details}

We primarily use ROUGE scores~\cite{lin2004rouge} to evaluate all trained models to compare the best distilled model against baseline methods. Since our work focuses on token-level distillation, we compare \OURS primarily with other token-level approaches. For evaluation, we use the non-overlapping test splits from each of the two datasets.

\paragraph{Baselines.} We evaluate \OURS against the following baselines:

\begin{itemize}[leftmargin=*]
\item \textbf{Supervised finetuning (SFT).} The model is directly finetuned on the training dataset without any distillation.
\item \textbf{Token-level distillation methods.} We compare \OURS with several KL-divergence-based token-level distillation approaches, including Forward KL (FKL)~\cite{hinton2015distilling}, Reverse KL (RKL). In the future, we'll add Skewed KL (SKL), and Adaptive KL (AKL)~\cite{wu2024rethinking}.

\item \textbf{Sequence-level distillation methods.} Although our method focuses on token-level distillation, we also include comparisons with state-of-the-art sequence-level distillation methods, namely SeqKD~\cite{kim2016sequence} and MiniLLM~\cite{gu2023minillm}. SeqKD can be seen as the sequence level forward KL distillation, while MiniLLM can be seen as the sequence level reverse KL distillation.
\end{itemize}

\subsection{Main Results}
We present the teacher model evaluation results in Table~\ref{table:teacher-results} and main results in Table~\ref{table:main-results}. 

\begin{table}[ht]
\centering
\resizebox{0.51\textwidth}{!}{
\begin{tabular}{l|c|ccc}
    \toprule
    \textbf{Benchmark} & \textbf{Method} & \textbf{ROUGE\_1} & \textbf{ROUGE\_2} & \textbf{ROUGE\_L} \\
    \midrule
    \multirow{6}{*}{Ultrachat} 
        & SFT & 0.4049 & 0.183 & 0.2478 \\
        & FKL & 0.5146 & 0.2477 & 0.3325 \\
        & RKL & 0.5070 & 0.2397 & 0.3301 \\
        & SeqKD & 0.5150 & 0.2502 & 0.3348 \\
        & MiniLLM & 0.5211& 0.2491 & 0.3347 \\
        & \textbf{\OURS} & \textbf{0.5407} & \textbf{0.2599} & \textbf{0.3425} \\
    \midrule
    \multirow{6}{*}{OpenMath}
        & SFT & 0.4531 & 0.2137 & 0.2305 \\
        & FKL & 0.4768 & 0.2327 & 0.2386 \\
        & RKL & 0.4757 & 0.2314 & 0.2377 \\
        & SeqKD & 0.4829 & 0.2389 & 0.2482 \\
        & MiniLLM & 0.4197 & 0.1843 & 0.2240 \\
        & \textbf{\OURS} & \textbf{0.5125} & \textbf{0.2664} & \textbf{0.2465} \\
    \bottomrule
\end{tabular}
}
\caption{Main results. Ultrachat refers to \texttt{ultrachat-200k}, OpenMath refers to the \texttt{OpenMathReasoning}.}
\label{table:main-results}
\end{table}

\begin{table}[ht]
\centering
\resizebox{0.51\textwidth}{!}{
\begin{tabular}{l|c|ccc}
    \toprule
    \textbf{Dataset} & \textbf{Base Model} & \textbf{ROUGE\_1} & \textbf{ROUGE\_2} & \textbf{ROUGE\_L} \\
    \midrule
    Ultrachat & Qwen2.5-7B-Instruct & 0.5726 & 0.2974 & 0.3707 \\
    \midrule
    OpenMath & Qwen2.5-7B-Instruct & 0.578 & 0.3451 & 0.2813 \\
    \bottomrule
\end{tabular}
}
\caption{Teacher model evaluation results. Ultrachat refers to \texttt{ultrachat-200k}, OpenMath refers to the \texttt{OpenMathReasoning}.}
\label{table:teacher-results}
\end{table}


As shown in Table~\ref{table:main-results}, our proposed \OURS method consistently outperforms all strong baselines on both Ultrachat and OpenMathReasoning. On ultrachat, \OURS achieves a Rouge‑1 of 0.5407, representing an absolute gain of +0.0261 over the next‑best FKL (0.5146) and +0.1358 over naive SFT (0.4049). We see similar improvements in Rouge‑2 (0.2599 vs. 0.2477 for FKL) and Rouge‑L (0.3425 vs. 0.3325 for FKL), indicating that \OURS not only captures richer bi‑gram patterns but also yields more coherent long‑form outputs. On the OpenMathReasoning  benchmark, \OURS attains Rouge‑1 score of 0.5125, outperforming FKL by +0.0357 and SFT by +0.0594. Its gains in Rouge‑2 and Rouge‑L further demonstrate robust generalization on complex reasoning tasks. These consistent, multi‑metric improvements validate that our unified distillation framework effectively transfers both the teacher’s generative knowledge and task‑specific classification capabilities into a compact student model, making it particularly well‑suited for real‑time candidate evaluation scenarios.


\section{Analysis}

\begin{figure}[t]
\centering
\includegraphics[width=1\linewidth]{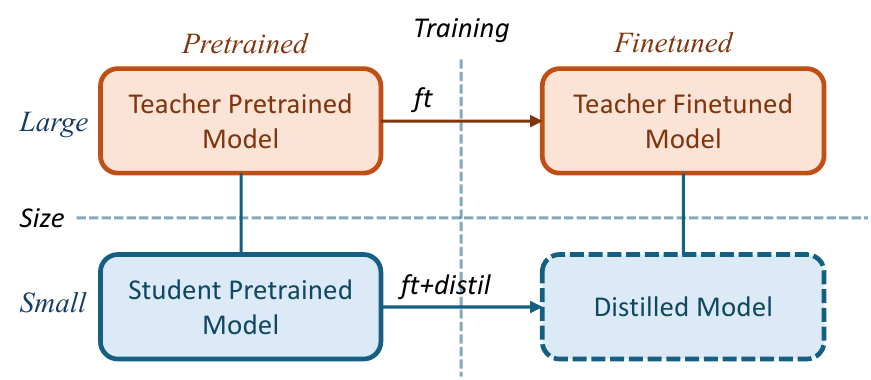 }
\caption{Parallel losses.}
\label{example_abs}
\vspace{-8pt}
\end{figure} 

Despite the main results, we also tested some different versions of \OURS.
Token-level knowledge distillation consists of three components: the teacher logit distribution, the student logit distribution, and a divergence function. Since the divergence function is typically fixed, the primary distinction arises from the two distributions, which the training objective seeks to align.
In the previous sections, we present \OURS as minimizing the divergence between the synthetic teacher distribution $\pi_s^*$, which comes from the distribution shift adjusted teacher distribution, and a student distribution $\pi_\theta$.
However, in this section, we extend Eq.~\ref{eq:delta-eq-8} to a more general setting. In this formulation, the synthetic teacher distribution $\pi_s^*$ is obtained by combining a base distribution with a distribution-shift term, whereas the student distribution $\pi_\theta$ is derived from the corresponding adjusted distribution.
Specifically, the loss functions we evaluate are summarized in Table~\ref{table:analysis-eq}, and the corresponding experimental results are reported in Table~\ref{table:analysis}. During our exploration of different parallel \OURS variants, we observed that some configurations are not tunable: their gradient norms during backpropagation grow excessively large, eventually causing training to diverge. We indicate these cases as “not tunable” in the last column. Notably, all non-tunable variants place the gradient inside the distribution shift term, preventing effective backpropagation across such a long dependency. This ultimately leads to unstable updates and causes training to diverge.

\begin{table}[ht]
\centering
\resizebox{0.51\textwidth}{!}{
\begin{tabular}{c|cccc}
    \toprule
    \textbf{Method} & \textbf{Stu. dist.} & \textbf{Tea. starting dist.} & \textbf{Dist. shift term} & \textbf{Tunable?} \\
    \midrule
    $V_1$ & $\pi_{\text{3B-sft}}$ & $\pi_{\text{3B-ref}}$ & $\text{diff}(\pi_{\text{7B-sft}}-\pi_{\text{7B-ref}})$ & \ding{51} \\
    $V_2$ & $\pi_{\text{3B-sft}}$ & $\pi_{\text{7B-sft}}$ & $\text{diff}(\pi_{\text{3B-ref}}-\pi_{\text{7B-ref}})$ & \ding{51} \\
    $V_3$ & $\pi_{\text{7B-sft}}$ & $\pi_{\text{7B-ref}}$ & $\text{diff}(\pi_{\text{3B-sft}}-\pi_{\text{3B-ref}})$ & \ding{55} \\
    $V_4$ & $\pi_{\text{7B-sft}}$ & $\pi_{\text{3B-sft}}$ & $\text{diff}(\pi_{\text{7B-ref}}-\pi_{\text{3B-ref}})$ & \ding{51} \\
    $V_5$ & $\pi_{\text{3B-ref}}$ & $\pi_{\text{7B-ref}}$ & $\text{diff}(\pi_{\text{3B-sft}}-\pi_{\text{7B-sft}})$ & \ding{55} \\
    $V_6$ & $\pi_{\text{3B-ref}}$ & $\pi_{\text{3B-sft}}$ & $\text{diff}(\pi_{\text{7B-ref}}-\pi_{\text{7B-sft}})$ & \ding{51} \\
    $V_7$ & $\pi_{\text{7B-ref}}$ & $\pi_{\text{7B-sft}}$ & $\text{diff}(\pi_{\text{3B-ref}}-\pi_{\text{3B-sft}})$ & \ding{55} \\
    $V_8$ & $\pi_{\text{7B-ref}}$ & $\pi_{\text{3B-ref}}$ & $\text{diff}(\pi_{\text{7B-sft}}-\pi_{\text{3B-sft}})$ & \ding{55} \\
    \bottomrule
\end{tabular}
}
\caption{\textit{Parallel} \OURS definitions. Stu. dist. refers to student distribution, and Tea. strating dist. refers to teacher model starting point distribution.}
\label{table:analysis-eq}
\end{table}

\begin{table}[ht]
\centering
\resizebox{0.51\textwidth}{!}{
\begin{tabular}{l|c|ccc}
    \toprule
    \textbf{Benchmark} & \textbf{Method} & \textbf{ROUGE\_1} & \textbf{ROUGE\_2} & \textbf{ROUGE\_L} \\
    \midrule
    \multirow{4}{*}{Ultrachat} 
        & $V_1$ & 0.5407 & 0.2599 & 0.3425 \\
        & $V_2$ & 0.5308 & 0.2539 & 0.3360 \\
        & $V_4$ & 0.5402 & 0.2604 & 0.3428 \\
        & $V_6$ & 0.5417 & 0.2604 & 0.3432 \\
    \bottomrule
\end{tabular}
}
\caption{\textit{Parallel} \OURS results.}
\label{table:analysis}
\end{table}

\section{Conclusion}
In this work, we proposed \OURS, a novel extension of token-level knowledge distillation that explicitly incorporates the distributional shift induced by teacher supervised finetuning. Unlike prior approaches that directly align student and teacher logits under the assumption of a shared representation space, \OURS constructs a synthetic target distribution that preserves both pretraining and finetuning signals while remaining compatible with the student’s representation space. Extensive experiments across instruction-following and reasoning datasets demonstrate that \OURS consistently improves student performance over strong KD baselines. Furthermore, our framework introduces a flexible formulation, including the Parallelogram Loss, that enables effective integration of multiple teacher variants. These results highlight the importance of modeling distributional shifts in KD and open avenues for more robust and generalizable distillation strategies for large language models.

\section*{Limitations}
\paragraph{Limitations on training.} Our current training architecture relies on CPU-based shared memory and PCIe communication between  processes. While simple and effective, this introduces latency due to GPU-to-CPU data transfers. A more efficient solution would leverage CUDA IPC and NVLink to enable direct GPU-to-GPU communication, eliminating communication bottlenecks and further improving training throughput.

\bibliography{custom}





\end{document}